\documentclass[format=sigconf]{acmart}
\usepackage[utf8]{inputenc}
\usepackage[ruled,vlined]{algorithm2e}
\usepackage{amsmath}
\usepackage{graphicx}
\usepackage{float}
\usepackage{pgf, tikz}
\usepackage{float}
\usepackage{algorithmic}
\usepackage{subcaption}
\usetikzlibrary{arrows, automata,positioning}
\usepackage{appendix}

\newcommand\independent{\protect\mathpalette{\protect\independenT}{\perp}}
\def\independenT#1#2{\mathrel{\rlap{$#1#2$}\mkern2mu{#1#2}}}

\newtheorem*{theorem*}{Theorem}
\newcommand{\ldata}{{\textbf{\textsc{L-data}}}}
\newcommand{\rdata}{{\textbf{\textsc{R-data}}}}
\title[Causal Transfer Random Forests]{Causal Transfer Random Forest: Combining Logged Data and Randomized Experiments for Robust Prediction}

\setcopyright{acmcopyright}
\copyrightyear{2021}
\acmYear{2021}
\setcopyright{acmcopyright}
\acmConference[WSDM '21] {Proceedings of the Fourteenth ACM International Conference on Web Search and Data Mining}{March 8--12, 2021}{Virtual Event, Israel}
\acmBooktitle{Proceedings of the Fourteenth ACM International Conference on Web Search and Data Mining (WSDM '21), March 8--12, 2021, Virtual Event, Israel}
\acmPrice{15.00}
\acmDOI{10.1145/3437963.3441722}
\acmISBN{978-1-4503-8297-7/21/03}

\author{Shuxi Zeng}
\affiliation{\institution{Duke University}}
\email{zengshx777@gmail.com}

\author{Murat Ali Bayir}
\affiliation{\institution{Microsoft}}
\email{muratmetu@yahoo.com}

\author{Joseph J. Pfeiffer III}
\affiliation{\institution{Microsoft}}
\email{joelpf@microsoft.com}

\author{Denis Charles}
\affiliation{\institution{Microsoft}}
\email{cdx@microsoft.com}

\author{Emre K\i c\i man}
\affiliation{\institution{Microsoft}}
\email{emrek@microsoft.com}

\begin{document}
\fancyhead{}
\begin{abstract}

It is often critical for prediction models to be robust to distributional shifts between training and testing data.
From a causal perspective, the challenge is to distinguish the stable causal relationships from the unstable spurious correlations across shifts.
We describe a {\em causal transfer random forest} (CTRF) that combines existing training data with a small amount of data from a randomized experiment to train a model which is robust to the feature shifts and therefore transfers to a new targeting distribution. Theoretically, we justify the robustness of the approach against feature shifts with the knowledge from causal learning.
Empirically, we evaluate the CTRF using both synthetic data experiments and real-world experiments in the Bing Ads platform, including a click prediction task and in the context of an end-to-end counterfactual optimization system. The proposed CTRF produces robust predictions and outperforms most baseline methods compared in the presence of feature shifts.
\end{abstract}



\maketitle


\section{Introduction} 
%
A central assumption of the majority of machine learning algorithms is that training and testing data is collected independently and identically from an underlying distribution.
Contrary to this assumption, in many scenarios training data is collected under different conditions than the deployed  environment~\cite{quionero2009dataset}.
For example, online services commonly use counterfactual models of user behavior to evaluate system and policy changes prior to online deployment~\cite{bayir2019genie}.  
In these scenarios, models train on interaction data gathered from previously deployed versions of the system, yet must make predictions in the context of the new system (prior to deployment).
Other domains with distribution or covariate shifts include text and image classification 
~\cite{daume2006domain,wang2018deep}, information extraction~\cite{ben2007analysis}, as well as prediction and now-casting~\cite{lazer2014parable}.  

\begin{figure}[h]
    \centering
    \includegraphics[scale=1.0]{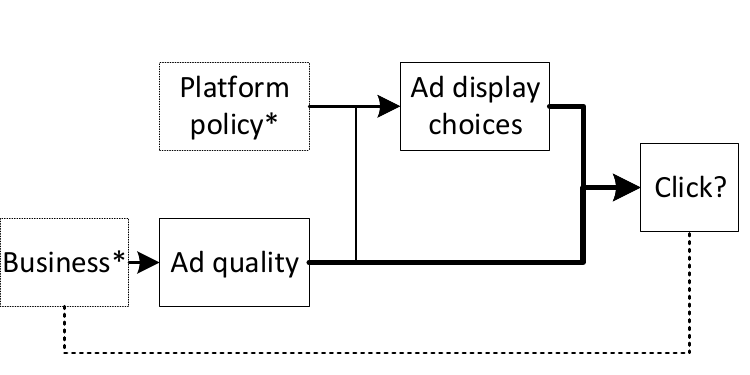}
    \caption{Challenges of robust prediction in a click prediction task:  While click likelihood depends on display choices and ad quality, those two factors will co-vary in a way that changes as platform policy shifts.  Other correlations (e.g,. business attributes) are unstable across environments.
    }
   \label{fig:adexample}
\end{figure}

Conventional machine learning algorithms exploit all correlations to predict a target value.  Many of these correlations, however, can shift when parts of the environment are unrelated to our task change.
Viewed from a causal perspective, the challenge is to distinguish causal relationships from unstable spurious correlations, as well as to disentangle the influence of co-varying features with the target value \cite{peters2016causal,rojas2018invariant,arjovsky2019invariant}.
For example, in the counterfactual click prediction task we may wish to predict whether a user would have clicked on a link if we change the page layout 
(Figure~\ref{fig:adexample}).  Training a prediction model based on current click logged data will find many factors related to an observation of a click ({\em e.g.}, display choices such as location and formatting, as well as factors related to ad quality and relevance).  Yet, these factors are often entangled and co-vary due to platform policy, such as giving higher quality links more visual prominence through their location and formatting.  In other cases, correlations may be unstable across environments as data generating mechanisms or the platform policy changes.
A click prediction model based on this data may be unable to determine how much the likelihood of a click is due to relevant contextual features versus environmental factors. As long as the correlations among these features do not change, the prediction model will perform well.  However, when the system is changed---perhaps a new page layout algorithm reassigns prominence or locations for links ---the prediction model will fail to generalize. Moreover, such drastic system changes are very common in practice, which will be discussed in the real-application section.

One way to disentangle causal relationships from merely correlational ones is through experimentation \cite{cook2002experimental,kallus2018removing}.  For example, if we randomize the location of links on a page it will break the spurious correlations between page location and all other factors.  This allows us to determine the true influence or the ``causal effect" of page location on click likelihood.  Unfortunately, randomizing all important aspects of a system and policy is often prohibitively expensive, as employing the random platform policy in the system generally induces revenue loss compared with the a well-tuned production system. Gathering the scale of randomization data necessary for building a good prediction model is frequently not possible.  Therefore, it is desirable to efficiently combine the relatively small scale randomized data and the large scale logged data for robust predictions after the policy changes.

In this paper, motivated by an offline evaluation application in the sponsored search engine, we describe a {\em causal transfer random forest} (CTRF). The proposed CTRF combines existing large-scale training data from past logs (\ldata) with a small amount of data from a randomized experiment (\rdata) to better learn the causal relationships for robust predictions. It uses a two-stage learning approach.  First, we learn the CTRF tree structure from the \rdata.  This allows us to learn a decision structure that disentangles all the relevant randomized factors.  Second, we calibrate each node (such as calculating the click probability) of the CTRF with both the \ldata~ and the \rdata.  The calibration step allows us to achieve the high-precision predictions that are possible with large-scale data.  Further, we complement our intuitions with theoretical foundations, showing that the model structure training on randomized data should provide a robust prediction across covariate shifts.

Our contributions in this paper are 3-fold. Firstly, we introduce a new method for building robust prediction models that combine large-scale \ldata~ with a small amount of \rdata. Secondly, we provide a theoretical interpretation of the proposed method and its improved performance from the causal reasoning and invariant learning perspective. Lastly, we provide an empirical evaluation of the robustness improvements of this algorithm in both synthetic experiments and multiple experiments in a real-world, large-scale online system at Bing Ads. The Supplementary Material including reproducible code, experiment details and theorem proofs is provided in the GitHub repository :\url{https://github.com/zengshx777/CTRF}.

\section{Related Work}
\subsection{Off-policy Learning in Online Systems}
This work is motivated from the task of performing offline policy evaluation in the online system~\cite{bottou2013counterfactual,li2012unbiased}. Occasionally, we would like to know the outcome of performing an unexplored tuning in the current system, which is also known as the counterfactual outcome. For example, we are interested in the change in users click probability after modifying the auction mechanism in the online ads system \cite{varian2007position}. Sometimes, the modifications can be drastic from the previous policy. Instead of running the costly online A/B testing \cite{xu2015infrastructure}, some offline methods are frequently used to predict the counterfactual outcomes based on the existing logged data from the current system.
One novel solution is to build the model-based simulator. Specifically, we build the model simulating the users behaviour and measure the metrics change after implementing the proposed policy changes in the simulator \cite{bayir2019genie}. We usually train the user-simulator model on the \ldata~ generating under previous platform policy. As a result, the covariate shift problem happens if the proposed change is drastic. 


\subsection{Transfer Learning and Domain Adaptation}
The discrepancy across training (large scale logged data e.g.) and testing (data after policy change e.g.) distribution is a long-standing problem in the machine learning community. Classic supervised learning might suffer from the generalization problem when the training data has a different distribution with the data for testing, which is also referred to the covariate (or distribution or dataset) shift problem, or the domain adaptation task~\cite{quionero2009dataset,bickel2009discriminative,daume2006domain}. Specifically, the model learned on a training data (source domain) is not necessarily minimizing the loss on the testing distribution (target domain). This hampers the ability of the model to transfer from one distribution or domain to another one. 

Some researchers propose to correct for the difference through sample reweighting~\cite{neal2001annealed, shimodaira2000improving,huang2007correcting}. 
Ideally, we wish to weight each unit in the training set so that we can learn a model minimizing the loss averaged on the testing distribution after reweighting. However, this strand of approaches requires the knowledge of the testing distribution to estimate the density and is likely to fail when the testing distribution deviates a lot from the training distribution, with extreme values in the density ratio.
Another type of methods is feature based. Some approaches aim at learning the features or representations that have predictive power while remaining a similar marginal distribution across source and target domain \cite{zhang2013domain,ganin2016domain}. However, the balance on marginal distributions does not ensure a similar performance on the target domain. We need to justify the predictive performance for the learnt features on the target domain.

\subsection{Causality and Invariant Learning}
Recently, some methods adapt the idea from causal inference to define the transfer learning with assumptions on the causality relationship among the features \cite{peters2016causal,magliacane2018domain, rojas2018invariant,meinshausen2018causality,kuang2018stable,arjovsky2019invariant,huang2020causal}. Specifically, researchers paraphrase the transfer difficulty as the confounding problem in causal inference literature \cite{pearl2009causalitybook,imbens2015causalbook}. The reason for poor generalization performance is that the model is learning some  spurious correlation relationships on the source domain, which are not expected to hold on the target domain. The invariant features across the domains should be the direct causes of the outcome (suppose being not intervened), as the causality relationship is presumably to be stable across training and testing distribution \cite{pearl2009causal}. Our work focus on utilizing the \rdata~ generating from a random policy, which is formally defined later, to exploit the causal relationship with limited sample size. Within the same causality framework, our model learns the invariant features that can transfer to the unknown target domain and be robust to severe covariate shifts.



\section{Causal Transfer Random Forest}




\subsection{Problem Setup}
Let $y\in \mathcal{Y}$ be a binary outcome label 
given contextual features $x\in \mathcal{R}^{p}$ and intervenable features, $z\in \mathcal{R}^{p'}$. 
We desire a model to map from the feature space to a distribution over the outcome space, {\em i.e.} learning the conditional distribution $p(y|x,z)$. Taking our motivating application, sponsored search, as a concrete example, the contextual features $x$ include user context and the query issued by the user; the features $z$ encode aspects that the publishers can manipulate, for instance, the location or the quality of the ads; and $y$ is whether or not a user clicked on the ad.  In practice, an advertising system takes many steps to create the pages showing the ads.

The feature shift problem arises when there is a drastic change in the joint features  distribution of $p(x,z)$. This shift might happen if the marginal distribution of contextual feature $p(x)$ varies. More commonly, the shift occurs when $p(z|x)$ changes to another distribution $p^{\ast}(z|x)$, namely, we change the data generating mechanism for $z$. This can happen when the platform policy change in the sponsored search system. In this case, the model learned from the training distribution $p(x,z)=p(x)p(z|x)$ might not generalize to the new distribution $p^{\ast}(x,z)=p(x)p^{\ast}(z|x)$. Therefore, we wish to learn a model $p(y|x,z)$ that is robust to the feature distribution, which can be safely transferred from original feature distribution $p(x,z)$ to the new $p^{\ast}(x,z)$.

We factorize the data $(x,z,y)$ in the following way\cite{bottou2013counterfactual}:
\begin{eqnarray}
\label{eq:factorize}
p(x,z,y)=p(x)p(z|x)p(y|x,z),
\end{eqnarray}
where $p(x)$ denotes the distribution of contextual variable, $p(z|x)$ represents how the platform manipulates certain features, such as the process of selecting ads and allocating each ad to the position on a page, which involves a complicated system including auction, filtering and ranking decisions \cite{varian2007position}. Here $p(y|x,z)$ is the user click model. One question of interest is how the click through rate $E(y)$ changes if we make modifications to the system, {\em i.e.}, replacing the usual mechanism $p(z|x)$ with a new one $p^{\ast}(z|x)$,
\begin{eqnarray}
\label{eq:predict_target}
E^{\ast}(y)=\int\int\int p(x)p^{\ast}(z|x)p(y|x,z)\textup{d}x\textup{d}z.
\end{eqnarray}
Feature shifts happen if some radical modifications are proposed, namely $p(z|x)$ differs significantly from $p^{\ast}(z|x)$. The user click model $p(y|x,z)$ cannot produce a reliable estimate for the new click through rate $E^{\ast}(y)$ as we usually learn the click model based on $p(x,z)$ while the testing data for prediction is drawn from $p^{\ast}(x,z)$. As $z$ depends on $x$ differently under various policies, the correlation between $z$ and $y$ might change after policy changes from $p(z|x)$ to $p^{\ast}(z|x)$. In such a scenario, we wish to build a model that can transfer from training distribution $p(x,z)$ to the target distribution $p^{\ast}(x,z)$, allowing one to evaluate the impact of radical policy changes.  

Currently, some publishers run experiments to randomize the features like the layout and advertisement in each impression shown to the user, which makes $z$ independent of $x$. Now, we formally define the \rdata~ as the data generated from $p(x)p(z)$, usually limited in size due to the low performance and revenue of a random policy. Meanwhile, we possess a large amount of past log data from the distribution $p(x)p(z|x)$, which we call \ldata. This leads to the opportunity to more efficiently use \rdata~ by pooling it with  large-scale \ldata. 

Although our approach is motivated by the online advertising setting, it is not restricted to this domain or binary classification task. 
We aim at building a robust model $p(y|x,z)$ transferring from the smaller \rdata~ and the large scale \ldata~ to the targeting source $p^{\ast}(x,z)$. We focus on the case that $p^{\ast}(x,z)$ differs drastically from $p(x,z)$, which is either due to the change in the policy $p(z|x)$ or the variation in contextual features $p(x)$. Although in this application, we may know $p^{\ast}(x,z)$ in advance, the proposed method does not require any prior knowledge on the density of targeting source. 
\vspace{-1em}
\subsection{Proposed Algorithm}

We base our algorithm on the random forest method \cite{breiman2001random}, adapting prior work on the honesty principle for building causal trees and forests~\cite{athey2016recursive,wager2018estimation}. Usually, the tree-based method is composed of two stages~\cite{hastie2005elements}: building decision boundary and calibrating each leaf value at the end of the branch to  produce an estimate $p_{i}$. Furthermore, the random forest framework performs bagging on the training data and building decision tree on each bootstrap data to reduce variance. Advantages of random forests include their simplicity and ability to be paralleled.

\begin{figure}[h]
	\centering
	\includegraphics[width=0.5\textwidth]{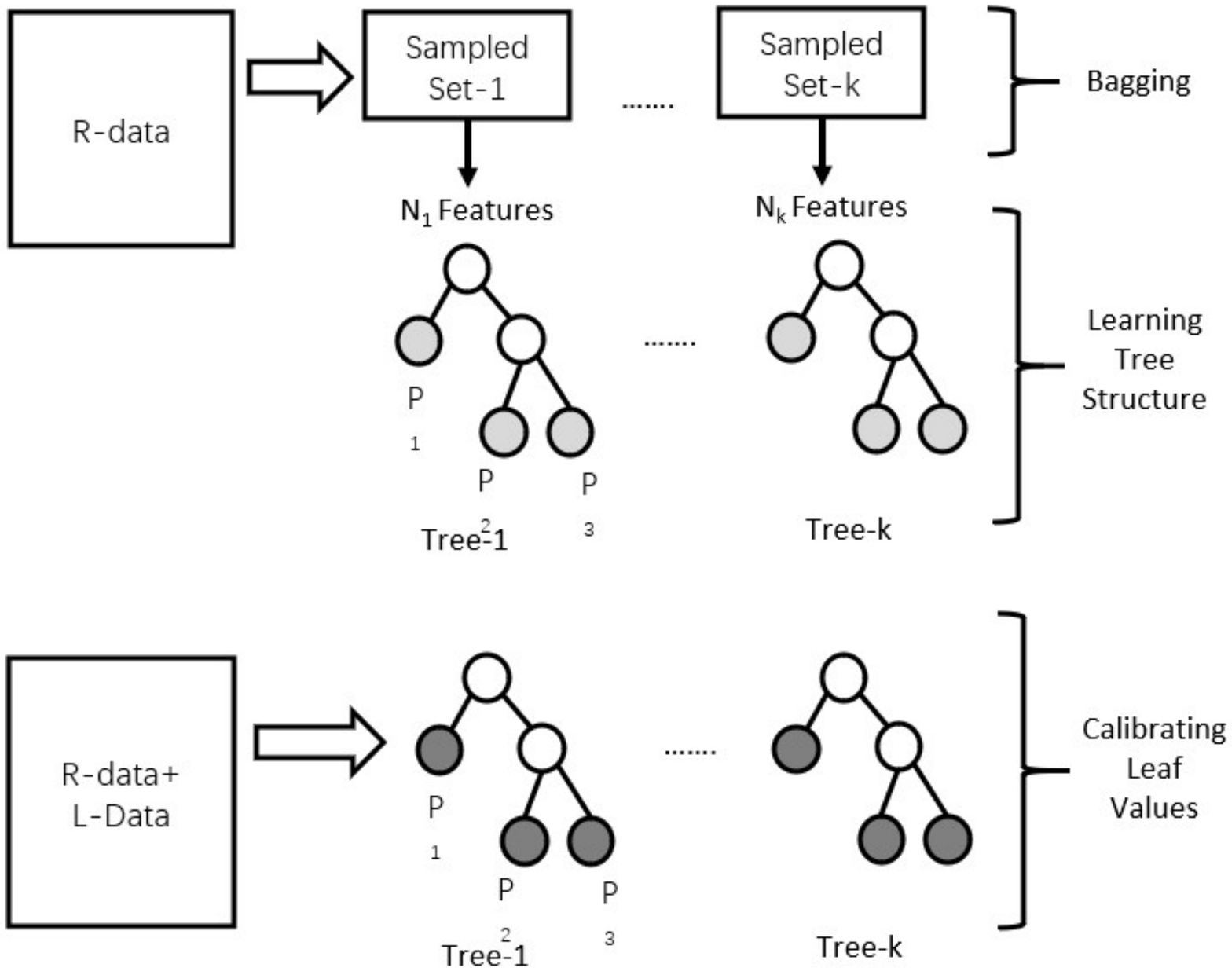}
	\caption{\small CTRF: building random forest from \rdata~ and \ldata \label{fig:framework}}
\end{figure}

To handle the feature shifts problem and use \rdata~ efficiently, we propose the Causal Transfer Random Forest (CTRF) algorithm. The framework is shown in Figure \ref{fig:framework}. We propose to do bagging and build decision trees solely on the \rdata~ and then calculate the predicted value ({\em e.g.}, click probability) on the nodes of each tree with pooled \rdata~ and \ldata. We make calibrations and aggregate over all trees with the simple average here, which can be extended to other approaches.
We describe the detailed algorithm in Algorithm \ref{alg:ctrf}.

We design the algorithm with the intuition that the \rdata~ reduces the problem of spurious correlation, one of the main reasons for the non-robustness of previous methods. Specifically, some of the correlations between $z$ and the outcome $y$ are influenced by the underlying generating mechanism, $p(z|x)$. In such cases, the correlation is spurious in the sense that it will disappear or change if we modify $p(z|x)$ to $p^{\ast}(z|x)$. The model trained on $p(x,z)$ will exploit those spurious correlations without the knowledge that the correlations will not hold on  distribution $p^{\ast}(x,z)$.  
It is important to note that the spurious and non-spurious components of $z$'s correlation with $y$ are often not well-aligned with the raw feature representation of $z$.  That is, this is not a feature selection problem.

\begin{algorithm}[h]
\caption{Causal Transfer Random Forest}
\label{alg:ctrf}
\scalebox{.8}{
\begin{minipage}{0.5\textwidth}
\begin{algorithmic}
   \STATE {\bfseries Input:} \rdata $\ \mathcal{D}^{R}=\{(x_{i},z_{i},y_{i}),i\in \mathcal{I}^{R}\}$, \ldata $\ \mathcal{D}^{L}=\{(x_{i},z_{i},y_{i}),i\in \mathcal{I}^{L}\}$ and the prediction point $(x^{\ast},z^{\ast})$.
   \STATE {\bfseries Hyperparameters:} bagging ratio: $r_{\textup{bag}}$; feature subsampling ratio: $r_{\textup{feature}}$; number of trees: $n^{\textup{tree}}$.\\
   \STATE {\bfseries Bagging:}	sample the data $\mathcal{D}^{R}$ with replacement for $n^{\textup{tree}}$ times with sampling ratio $r_{\textup{bag}}$ and sample on the feature set $(x,z)$ for each bootstrap data with ratio $r_{\textup{feature}}$.
   \FOR{$b=1$ {\bfseries to} $n^{\textup{tree}}$}
   \STATE {\bfseries{Learn decison tree}} For the bootstrapped data, $\{(x_{i}^{b},z_{i}^{b},y_{i}^{b})\}$, build decision tree $\mathcal{T}_{b}$ and corresponding leaf nodes $\mathcal{L}_{j}^{b}\subset \mathcal{R}^{p+p'} ,j=1,2,\cdots,L_{b}$, $L_{b}$ is the number of nodes for $\mathcal{T}_{b}$ by maximizing the Information Gain (IG) or Gini Score. 
\STATE{\bfseries{Calibrations}}
 For each node $\mathcal{L}_{j}^{b}$, we calculate the predicted value by the mean value of samples in this node: $\hat{y_{j}}^{b}=\bar{y_{i}},(x_{i},z_{i})\in \mathcal{L}_{j}^{b},i\in\mathcal{I}^{R}\cup\mathcal{I}^{L}$.
\ENDFOR 

\STATE{\bfseries {Predictions}} Collect the predicted value $\hat{y}^{b}$ for each $\mathcal{T}_{b}$ by examining the node that $(x^{\ast},z^{\ast})$ belongs and produce a prediction after aggregation, such as $\hat{y}=\bar{\hat{y}}^{b}$.

\STATE{\bfseries{Output}} Random forest $\{\mathcal{T}_{b},b=1,\cdots,n^{\textup{tree}}\}$ and prediction $\hat{y^{\ast}}$.
\end{algorithmic}
\end{minipage}%
}
\end{algorithm}

\begin{figure}[h]
    \centering
    \begin{tikzpicture}
    [
	> = stealth, 
	shorten > = 0.0pt, 
	auto,align=center,node distance= 0.5cm and 0.5cm,
	semithick 
	]
	\tikzstyle{every state}=[
	draw = white,
	thick,
	fill = white,
	minimum size = 3mm,
	]
	\node(A) (A){$x$: Ads relevance};
	\node[below left=0.2cm and 0.1 cm of A](B) {$z$: Position on the page};
	\node[below right=0.2cm and 0.1 cm of A](C) {$y$: Click or not click};
	\path[dashed,->] (A) edge node {} (B);
	\path[->] (A) edge node {} (C);
	\path[->] (B) edge node {} (C);
    \end{tikzpicture}
    \caption{Causal Directed Acyclic Graph (DAG) for the online advertisement system}
    \label{fig:causal_dag_ads}
\end{figure}
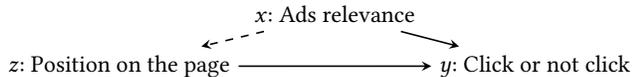

Figure \ref{fig:causal_dag_ads} demonstrates a spurious correlation instance in the ads system, depicting the relationships between ads relevance $x$, position $z$ and the click outcome $y$. The solid lines represent the ``stable'' relationship or effect between the ads relevance or the position and the click, while the dashed line stands for the relationship we can manipulate. In the \ldata, the position is not randomly assigned but instead associated with other features like ads relevance\cite{bottou2013counterfactual}. We tend to allocate ads of higher relevance to the top of the page. However, the correlation between position and click changes if we alter the policy allocating the position based on the relevance, namely $p(z|x)$.  
Despite the correlation between position and click being partially spurious, there is still a causal connection as well---higher positioned ads do attract more clicks, all else being equal.

Suppose the tree algorithm makes a split on the position feature, subsequently it becomes hard to detect the importance of relevance in two sub-branches split by position. As a result, if we only train on \ldata, the decision tree is likely to underestimate the importance of ad relevance. We wish the decision tree structure we learn to disentangle the unstable or spurious aspects of the correlation among the features and only learn the ``stable'' relationships. This task can be accomplished with the \rdata~ as it removes the spurious correlation. We formally define the ``stable'' relationship and prove why \rdata~ can learn those  relationships in the next section.

\subsection{Interpretations from Causal Learning}
In this section, we justify our intuitions in the previous sections theoretically based on the results in causal learning.
Previous literature builds the connections between the capability to generalize and the conditional invariant property. Theorem 4 in \cite{rojas2018invariant} demonstrates that if there is a subset of features $S^{\ast}$ that are conditionally invariant, namely the conditional distribution $y|S^{\ast}$ remains unchanged across  different distributions of $p(x,z,y)$, then the model built on those features $S^{\ast}$ with pooling data, $E(y_{i}|S_{i}^{\ast})$,  gives the most robust performance. The robustness is measured by the worst performance with respect to all possible choices of the targeting distribution $p(x,z)$, which further ensures the model can transfer.
This theorem indicates that we should build model on the set of features or the transformed features with conditional invariant property.

However, learning the stable features is not simple given we have only two types of distribution, 
The next theorem from \cite{peters2016causal,rojas2018invariant} states the relationship between conditional invariance and causality. Specifically, if we assume there are causal relationships or structural equation models (SEM) \cite{pearl2009causalitybook}, the direct causes of the outcome are the conditionally invariant features , $S^{\ast}=\textup{PA}_{Y}$, where $\textup{PA}_{Y}$ denotes the parents/direct causes for the outcome $y$.

With two well-established theorems above, we can look for the direct causes instead of the conditional invariant features. The following theorem shows that the \textbf{\textsc{R-data}} offers such opportunity.

\begin{theorem*}[Retain stable relationships with \rdata] 
\label{thm:random_est}
Assume $(x_{i},z_{i},y_{i})$ can be expressed with a direct acyclic graph (DAG) or structural equation model (SEM). 
Then the model trained on $\rdata$, $p(x_{i},z_{i})=p(x_{i}^{1})p(x_{i}^{2})\cdots p(x_{i}^{p})p(z_{i}^{1})p(z_{i}^{2})\cdots p(z_{i}^{p'})$ is consistent for the most robust prediction:
\begin{eqnarray}
\hat{E}(y_{i}|x_{i},z_{i})\Rightarrow E(y_{i}|\textbf{PA}^{Y})=E(y_{i}|S_{i}^{\ast})
\end{eqnarray}
\end{theorem*}
The theorem assumes all the variables $(x_{i},z_{i})$ are randomized and independent with each other in \rdata, which has a gap to the \rdata~ in practice as we cannot randomize the contextual features $x$. If the relationships between contextual features $x$ and outcome $y$ are unstable, it is hard to learn the stable relationships without randomizing on $x$. However, randomizing on the manipulable features $z$ will suffice in practice as the correlation between $x$ an $y$ is likely to be stable. For instance, the relationship between the user preference or the ads quality itself and the intention to click is expected to remain unchanged even if we switch the platform policy on displaying ads. The theorem above suggests if the model is trained on \rdata, it actually relies on the direct causes or robust features $S_{i}^{\ast}$ to make prediction. The detailed theorem proof is provided in the Supplementary Material.

Likewise, CTRF firstly learns the structure of the model or identifies the stable features for splitting the trees merely with the \rdata. With our random forest method, the stable features are the leaves sliced in the decision tree, which can be viewed as a transformation of the raw features. This step serves as an analogy to search for the direct causes or extract robust features. The calibration step on the leaf values with pooled data corresponds to make predictions conditioning on all robust features. The second step will not be ``contaminated'' by the spurious correlation in $\ldata$ as the the decision tree structure has already identified a valid adjustment set with \rdata~ and is conditioning on that. We also investigate whether the proposed method can pick up the stable features in the synthetic experiments to demonstrate its theoretical property.

\section{Experiments on Synthetic Data}
\subsection{Setup and Baselines}
In this part, we evaluate the proposed method and compare with several baseline methods in the presence of covariate shifts. Given it is a novel scenario (small amount of \rdata~ with large \ldata), we design two synthetic experiments to create an artificial case that the data generating mechanism $p(z|x)$ changes. The first experiment specifies the causality relationship between variables explicitly. The second experiment is a simulated auction similar to the real-world online, in which the relationship between variables are specified implicitly. In both experiments, we have some parameters controlling the degree of covariate shift which allows us to evaluate the performance against different degree of distributional variation.

In our experiments, we compare the {\em causal transfer random forest} (CTRF) with the following methods: logistic regression (LR) \cite{menard2002applied}, Gradient Boosting Decision Tree (GBDT) \cite{ke2017lightgbm}, logistic regression with sampling weighting (LR-IPW), Gradient Boosting Decision Tree with sample reweighting (GBDT-IPW), random forest model trained on \rdata~ (RND-RF), random forest model trained on \ldata~ (CNT-RF), random forest model trained with the \ldata~ and \rdata~ pooling together (Combine-RF). Among all those methods, LR-IPW and GBDT-IPW are designed to handle distribution shifts with a proper weighting with ratio of densities \cite{bickel2009discriminative, huang2007correcting}. Implementation details are included in the Supplementary Material.

As our method is designed to handle extreme covariate shifts, we evaluate different methods in terms of the performance on the shifted testing data only. Although our method is not restricted to classification task, we only focus on the binary outcome to be coherent with our motivated application from ads click. For binary classification task, we focus on the following two metrics, AUC (area under curve) and the cumulative prediction bias, $|\bar{\hat{y}}_{i}-\bar{y_{i}}|/\bar{y_{i}}$, which is the adjusted difference in the mean value of predicted values and actual outcomes. AUC captures the prediction power of the model while the cumulative prediction bias captures how our method can predict the counterfactual change, such as the change in the overall click rate.

\subsection{Synthetic Data with Explicit Mechanism}
\begin{figure}[h]
    \centering
\begin{subfigure}[b]{0.13\textwidth}\centering
    \begin{tikzpicture}
    [
	> = stealth, 
	shorten > = 0.0pt, 
	auto,align=center,node distance= 0.0cm and 
	0.0cm,
	semithick 
	]
	\tikzstyle{every state}=[
	draw = white,
	thick,
	fill = white,
	minimum size = 3mm,
	]
	\node(A) (A){$S$};
	\node[below left=0.6cm and 0.3cm of A](B) {$y$};
	\node[below right=0.6cm and 0.3cm of A](C) {$V$};
	\path[->] (A) edge node {} (B);
    \end{tikzpicture}
    \caption{$S\independent V$}
\end{subfigure}
\begin{subfigure}[b]{0.13\textwidth}\centering
    \begin{tikzpicture}
    [
	> = stealth, 
	shorten > = 0.0pt, 
	auto,align=center,node distance= 0.0cm and 
	0.0cm,
	semithick 
	]
	\tikzstyle{every state}=[
	draw = white,
	thick,
	fill = white,
	minimum size = 3mm,
	]
	\node(A) (A){$S$};
	\node[below left=0.6cm and 0.3cm of A](B) {$y$};
	\node[below right=0.6cm and 0.3cm of A](C) {$V$};
	\path[->] (A) edge node {} (B);
	\path[->] (A) edge node {} (C);
    \end{tikzpicture}
    \caption{$S\rightarrow V$}
\end{subfigure}
\begin{subfigure}[b]{0.13\textwidth}
\centering
    \begin{tikzpicture}
    [
	> = stealth, 
	shorten > = 0.0pt, 
	auto,align=center,node distance= 0.0cm and 
	0.0cm,
	semithick 
	]
	\tikzstyle{every state}=[
	draw = white,
	thick,
	fill = white,
	minimum size = 3mm,
	]
	\node(A) (A){$S$};
	\node[below left=0.6cm and 0.3cm of A](B) {$y$};
	\node[below right=0.6cm and 0.3cm of A](C) {$V$};
	\path[->] (A) edge node {} (B);
	\path[->] (C) edge node {} (A);
    \end{tikzpicture}
    \caption{$V\rightarrow S$}
\end{subfigure}
\caption{\label{fig:SV}Three possible relationships among the variables}
\end{figure}
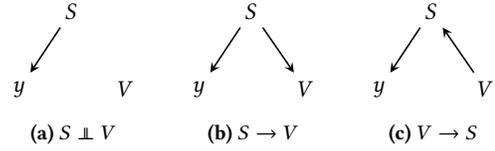

We generate the data in a similar fashion with the experiments in \cite{kuang2018stable}. We generate two sets of features $S,V$ for predictions. $S$ represents the stable feature or the direct cause of the outcome while $V$ represents the unstable factors that have spurious correlation with the outcome. We consider three possible scenarios for the relationships between $(S,V)$: (a)$S\independent V$, $S$ and $V$ are independent; (b) $S\rightarrow V$, $S$ is the cause for $V$; (c) $V\rightarrow S$, $V$ is the cause for $S$. Figure \ref{fig:SV} demonstrates these three cases. In all cases, $S=(S_{1},\cdots,S_{p_{s}})$ is the stable feature while $V=(V_{1},\cdots, V_{p_{v}})$ is the possible unstable factors sharing spurious correlation with the outcome.

In case (a), we generate $(S,V)$ from independent standard Normal distributions and transform them into the binary vectors,
\begin{gather*}
\tilde{S_{j}},\tilde{V_{k}}\sim \mathcal{N}(0,1),\ \ 
S_{j}=\textbf{1}_{\tilde{S}_{j}>0},V_{k}=\textbf{1}_{\tilde{V}_{k}>0}.
\end{gather*}
In case (b), we generate $S$ from Normal distributions first and generate $V$ as a function of $S$.
\begin{gather*}
\tilde{S}_{j}\sim \mathcal{N}(0,1), \tilde{V}_{k}=\tilde{S}_{k}+\tilde{S}_{k+1}+\mathcal{N}(0,2),\ \
S_{j}=\textbf{1}_{\tilde{S}_{j}>0},V_{k}=\textbf{1}_{\tilde{V}_{k}>0}.
\end{gather*}
In case (c), we generate $V$ first and simulate $S$ as a function of $V$.
\begin{gather*}
\tilde{V}_{k}\sim \mathcal{N}(0,1), \tilde{S}_{j}=\tilde{V}_{j}+\tilde{V}_{j+1}+\mathcal{N}(0,2),\ \
S_{j}=\textbf{1}_{\tilde{S}_{j}>0},V_{k}=\textbf{1}_{\tilde{V}_{k}>0}.
\end{gather*}
For the outcome, we keep the generating procedure same across three cases. The binary outcome $y$ is generated solely as a function of $S$,
\begin{gather*}
\tilde{y}=\textup{sigmoid}(\sum_{j=1}^{p_{s}}\alpha_{j}S_{j}+\sum_{j=1}^{p_{s}-1}\beta_{j}S_{j}S_{j+1})+\mathcal{N}(0,0.2),\ \
y=\textbf{1}_{\tilde{y}>0.5},
\end{gather*}
where $\textup{sigmoid}(x)=1/(1+\textup{exp}(-x))$. This specification includes both the linear and non-linear effect of $S$. The parameters take values as $\alpha_{j}=(-1)^{j}(j\%3+1)*p/3,\beta_{j}=p/2$. 

In addition to different generating mechanisms, we introduce an additional spurious correlation with biased sample selection. Specifically, we set an inclusion rate $r=(0,1)$ to create a spurious correlation between $y$ and $V$. If the average value of $\bar{V}_{i}=\sum_{j=1}^{p_{v}}V_{ij}$ and $\tilde{y}_{i}$ exceed or fall below 0.5 together, we include sample $i$ with probability $r$.  Otherwise, we include the sample with probability $1-r$. Namely, if $r>0.5$, $V$ and $y$ share positive correlation and the correlation is negative if $r<0.5$. The parameter $r$ controls the degree of spurious correlation which induces the covariate shifts. 

We generate a small amount of \rdata~ following case (a) with size $n_{r}=1000$, a large amount of \ldata~ following case (b) $n_{l}=5000$ and the testing data from case (c) with size $n_{t}=2000$ to mimic the policy change on testing data. We create a lower amount of \rdata~ to mimic the real business scenario that randomizing the platform policy reduces the revenue and thus being expensive to collect. We keep a slightly larger proportion of \rdata~ than the one in practice for fair comparisons (such as RND-RF) to demonstrate the essential advantage of the proposed method. Additionally, we set $r=0.7$ on the \ldata~ and let $r$ vary from $0.1$ to $0.9$ on the testing data to create additional deviance in the distribution. We also vary the number of features in total $p\in [20,40,80]$ and keep $p_{s}=0.4 p$. Within each configuration, we perform the experiments 200 times and calculate the average AUC and cumulative bias.

\begin{figure}[h]
    \centering
    \includegraphics[width=0.48\textwidth]{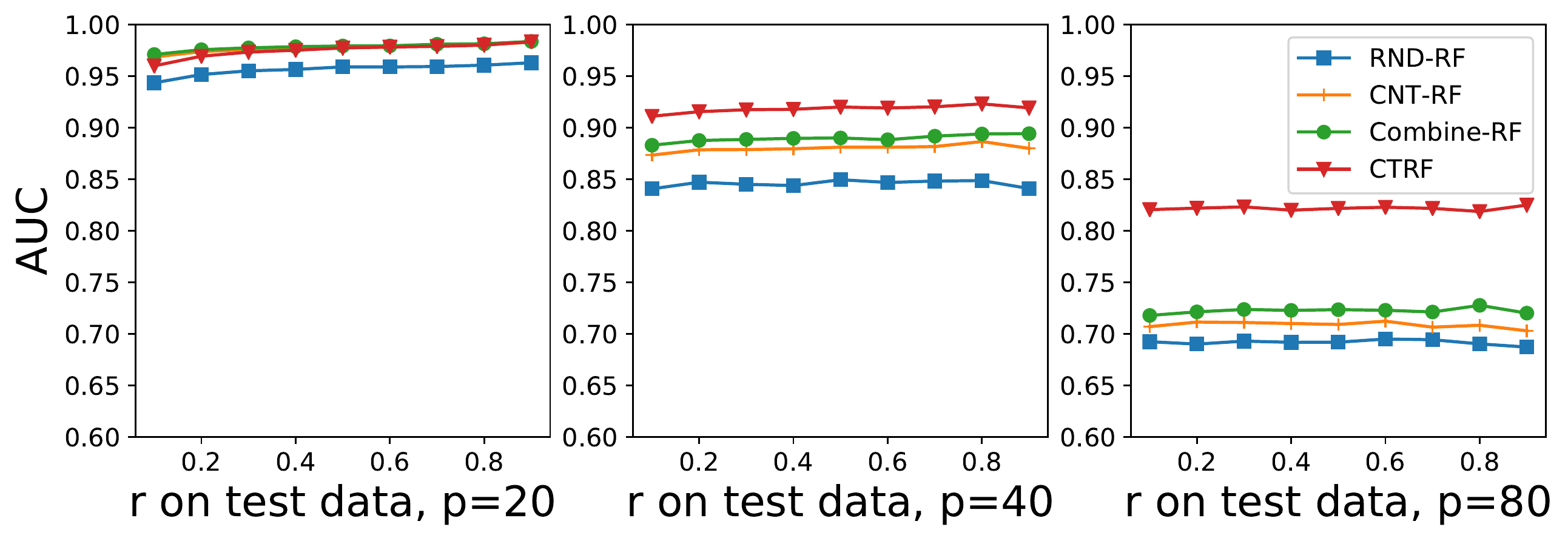}\\
    \includegraphics[width=0.48\textwidth]{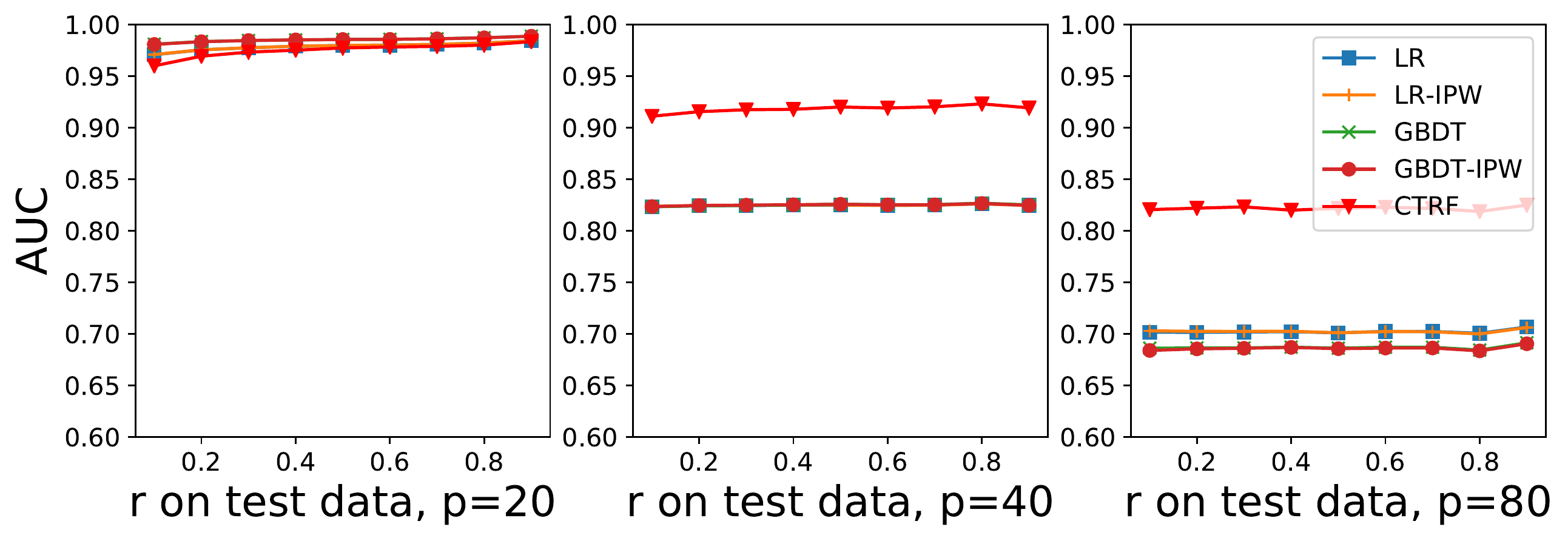}
    \caption{AUC comparison when $p=20,40,80$. The top row compares with random forest based method and the bottom row compares other baselines. CTRF produces largest AUC in most cases.}
     \label{fig:AUC_Compare}
\end{figure}

Figure \ref{fig:AUC_Compare} shows the comparison of AUC against the variation on both $p$ and $r$. The top row demonstrates the comparison within the domain of random forest. The CTRF (red lines) performs the best regardless of feature dimensions. The second row in Figure \ref{fig:AUC_Compare} shows the comparison with LR, LR-IPW, GBDT and GBDT-IPW. Although the performances are indistinguishable when $p=20$, the advantage of CTRF emerges as we have more spurious correlations.
\begin{figure}[h]
    \centering
    \includegraphics[width=0.48\textwidth]{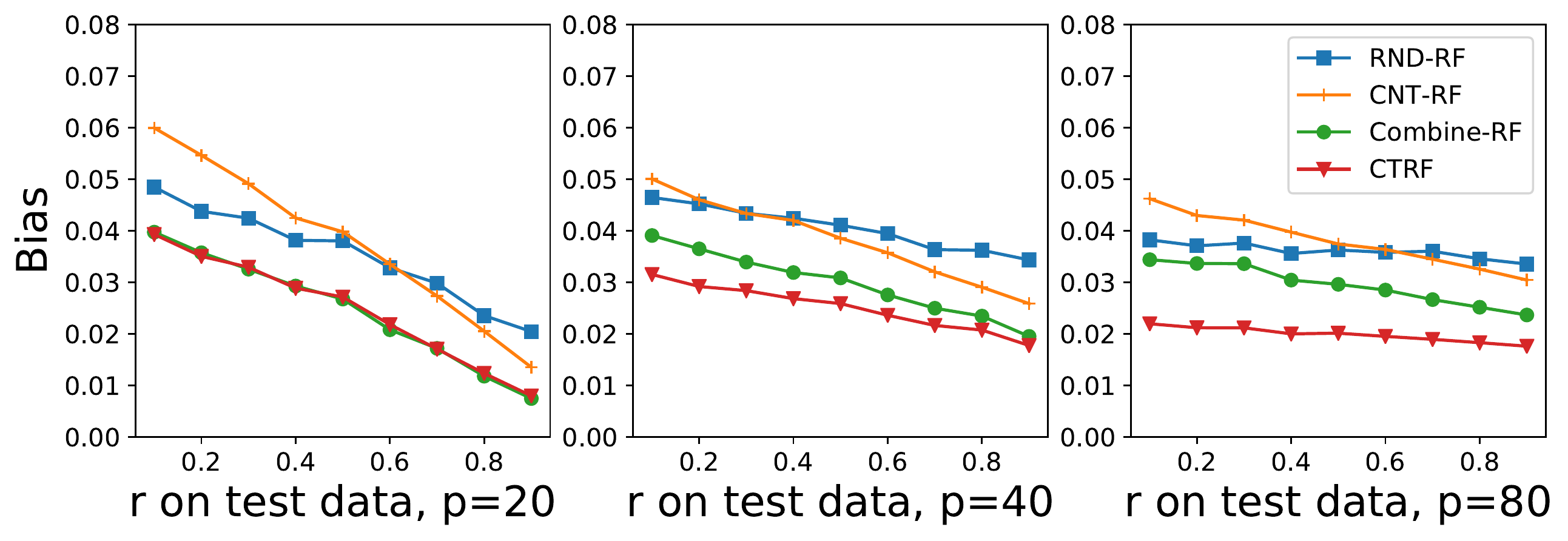}\\
    \includegraphics[width=0.48\textwidth]{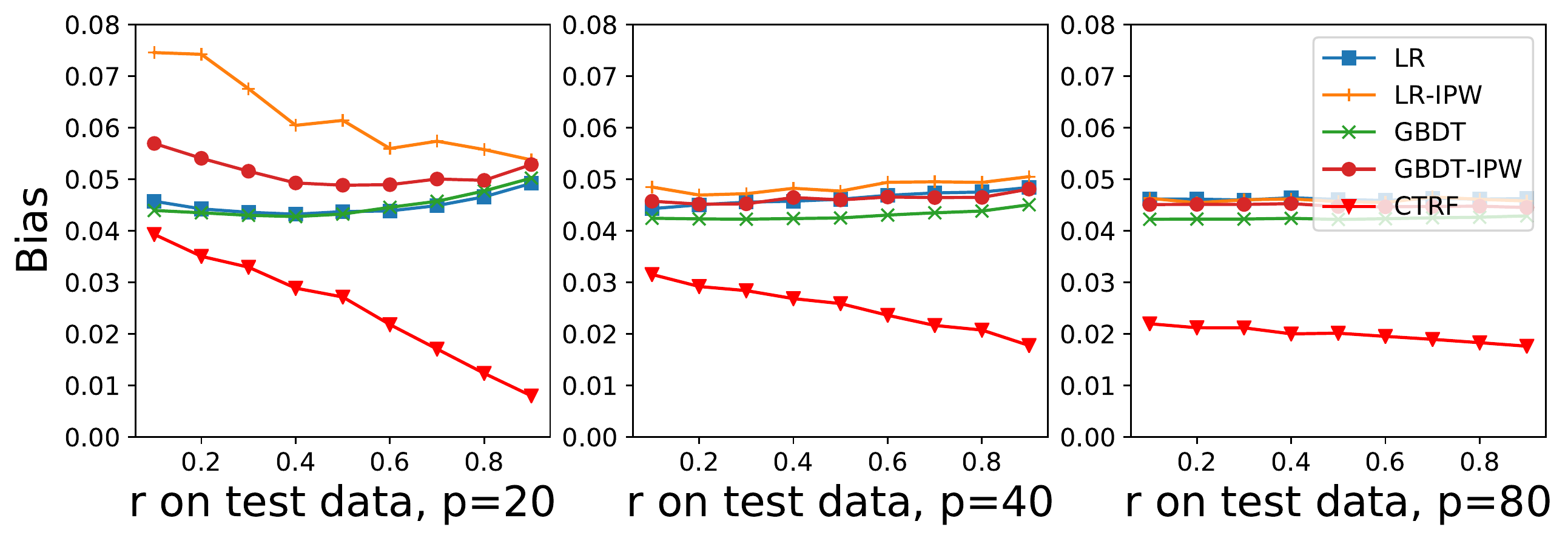}
    \caption{Bias comparison when $p=20,40,80$, with top row comparing with random forest based method and bottom row comparing other baselines. CTRF achieves the lowest bias in all cases.}
    \label{fig:Bias_Compare}
\end{figure}

Figure \ref{fig:Bias_Compare} shows the comparison in terms of the bias. A lower value represents a better performance. The top row shows the comparison with other random forest based methods. Generally, the cumulative bias increases as $r$ on the testing data decreases, which means the testing data deviates more from the \ldata. However, the advantage of CTRF (red lines) increases slightly as $r$ decreases, which demonstrates the robustness against covarites shifts.  The comparison with LR or GBDT based methods at the bottom row shows a similar trend with the AUC. The CTRF achieves a lower bias among all the approaches and its advantage increases as we have more features.

In terms of the scalability, we find that the advantage of CTRF over other methods increases as the feature size $p$ goes up, with a larger AUC and smaller bias. Additionally, the CTRF builds the decision tree solely on the \rdata~ and the calibration stage on the pooled data is much less computationally intensive, which further demonstrates its advantage in scalability.

\subsection{Synthetic Auction: Implicit Mechanism}

In this subsection, we setup a synthetic auction scenario with a single tuning parameter in the policy, demonstrating both how simple parameters can introduce bias into a domain and CTRF's ability to transfer between them. 
\begin{figure*}[h]
    \centering
     \includegraphics[width=0.31\textwidth]{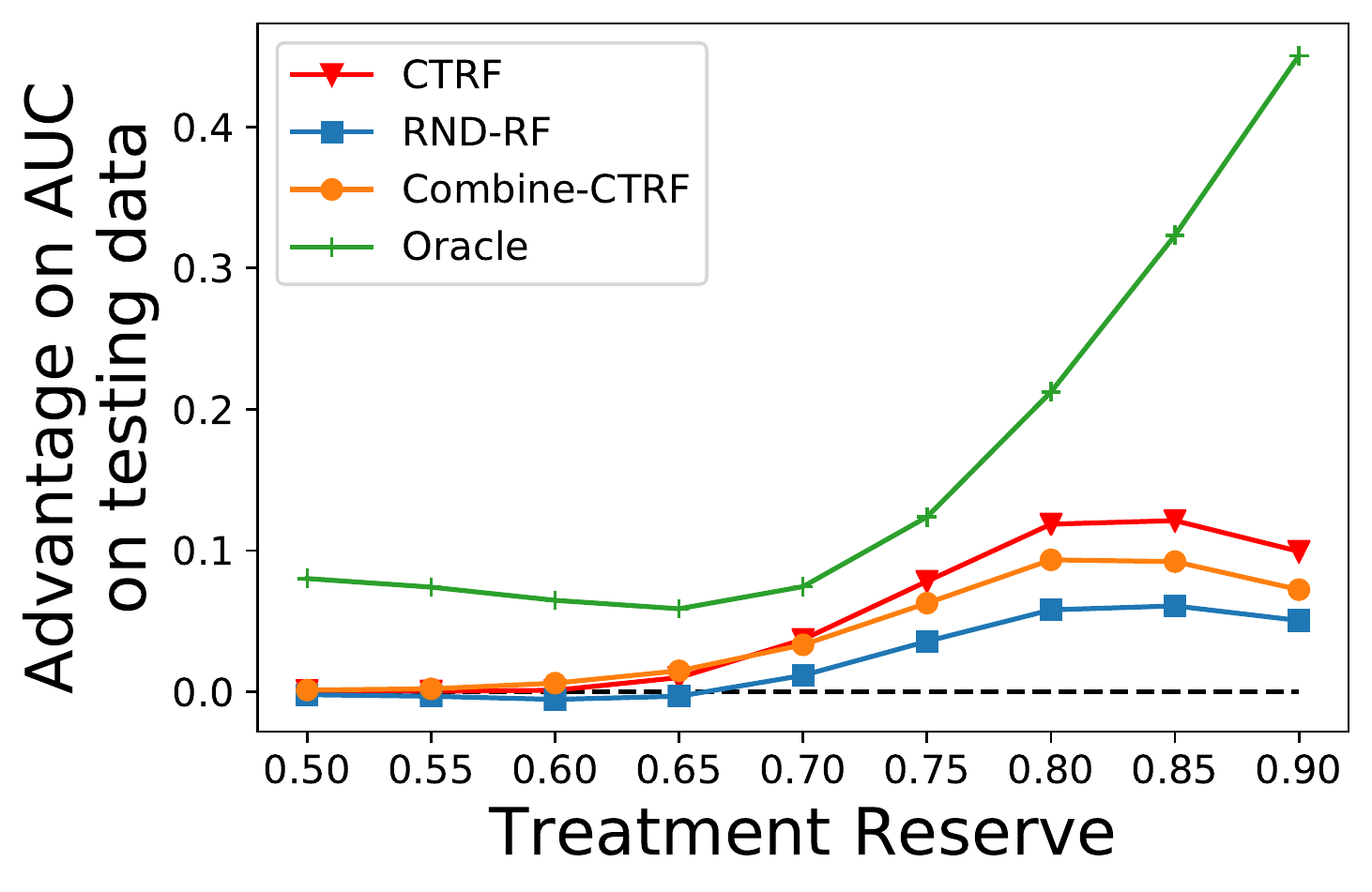}\hspace{1em}
    \includegraphics[width=0.31\textwidth]{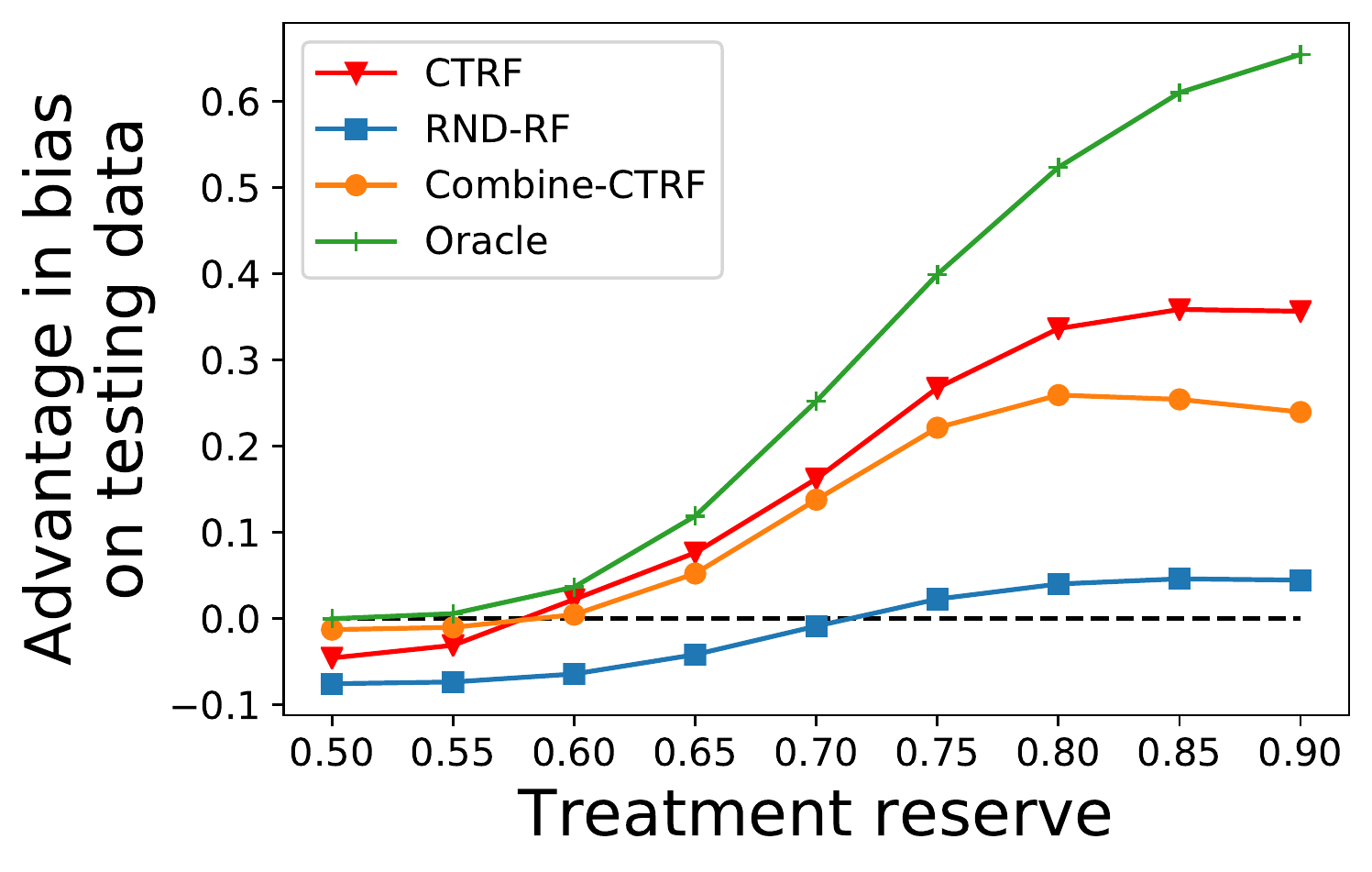}\hspace{1em}
    \includegraphics[width=0.31\textwidth]{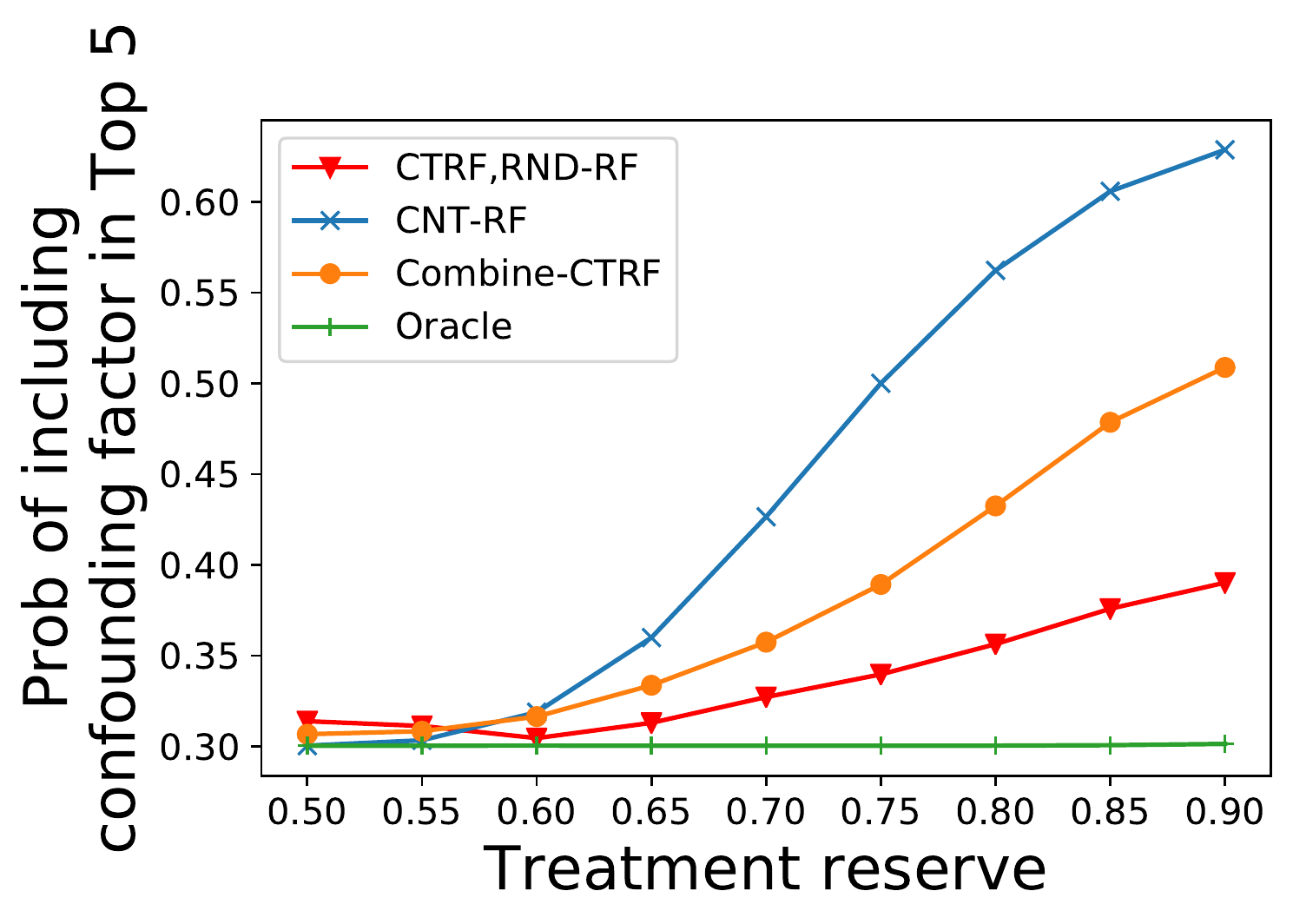}
    \caption{AUC (left graph), cumulative prediction bias (middle graph) and probability of including confounding factor "position" as Top 5 important features (right graph) versus treatment reserve $r$. Higher $r$ represents a larger change in the testing distribution. CTRF performs the best among all random forest methods.}
    \label{fig:auction_compare}
\end{figure*}
We first generate synthetic samples of classification data, or a mapping from features to a true relevant/irrelevant binary label.
From this data, we build a true relevance model with random forest to estimate the probability an item is relevant. Second, we build our \ldata~ and testing auctions by sampling (20 per auction) from the underlying relevance features and assigning a relevance score.  Per auction, the items are thresholded with the corresponding {\em relevance reserve} parameter and the remaining items are ranked.  This provides layout and position information, in addition to the relevance score and relevance features.  Third, Given the layout and items, a simulated user chooses a single ad as relevant uniformly at random to click, and leaves the others not clicked. The choice of click is uniform across positions, which means that \textit{position} is purely a factor spuriously correlated with the relevance while not affecting the click.
We provide the detailed generating mechanism in the Supplementary Material.
\begin{figure}[H]
    \centering
     \includegraphics[width=0.5\textwidth, trim={0cm 6.5cm 0cm 7.5cm}, clip=true]{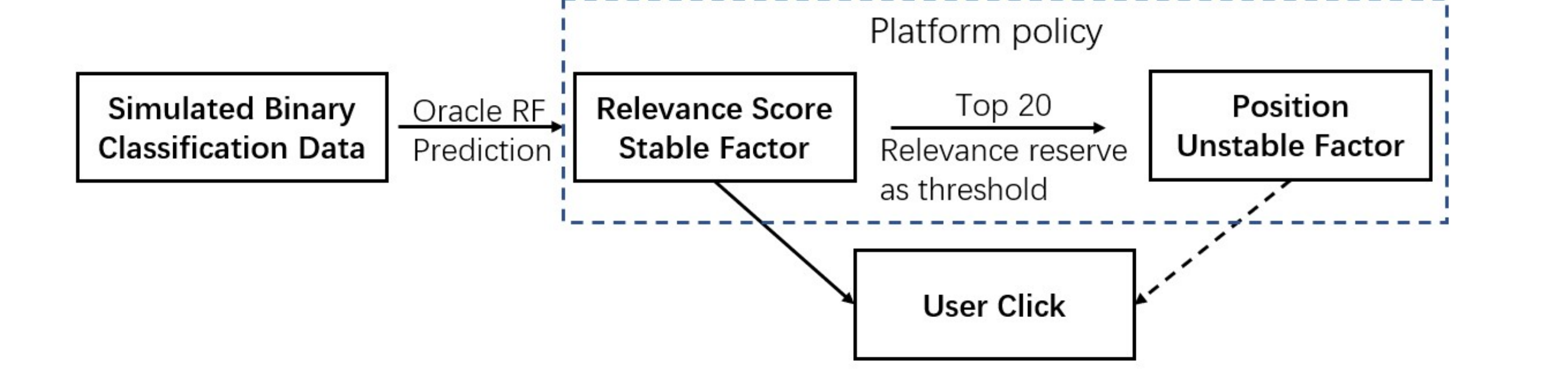}
    \caption{Procedures for simulating auctions. Position is an unstable factor for predicting click as the users pick ads uniformly on a page to click and its correlation with relevance score varies across policy, which is implicitly determined by the relevance reserve parameter.}
    \label{fig:auction_mechanism}
\end{figure}

The tuning parameter in the experiment is the {\it relevance reserve} parameter $r$, controlling the requirement that any item shown to a user meet a minimum relevance, which controls $p(z|x)$ implicitly. The mechanism to generate simulated auction is illustrated in Figure \ref{fig:auction_mechanism}. This parameter affects the correlation between relevance and position, which can vary between \ldata~ and testing data. Specifically, we generate the \ldata~ with relevance reserve parameter $r=0.5$ while the testing data with the relevance reserve varying in $r\in[0.5,0.9]$, simulating a desire to increase the quality of items presented to a user (with a higher threshold). A larger value in $r>0.5$ represents a higher deviation from the \ldata~ with $r=0.5$. For the \rdata, we do not have the auction procedure and we pick up the advertisement uniformly random to display on the page. The size of \rdata~ is approximately 20\% of the \ldata.

As we use the random forest model to generate the true relevance score, we compare the CTRF within the domain of random forest based methods only, including CNT-RF, RND-RF, Combine-RF and the oracle one training RF on the testing data. Figure \ref{fig:auction_compare} illustrates prediction performance of all method while setting CNT-RF as the baseline. To illustrate the advantage over the baseline method, CNT-RF, we minus the AUC of CNT-RF from that of all other methods and minus the bias of the corresponding model from the bias of CNT-RF. Therefore, a larger value in the graphs indicates a better performance of the corresponding method.

In Figure \ref{fig:auction_compare}, we observe that when the reserve for testing data lies close to $0.5$, all models show similar performance. However, as we increase $r$ on testing data and raise the degree of covariate shift, the CTRF method (red lines) greatly improves in both AUC and bias. Also, the CTRF demonstrates a  better prediction power and lower bias compared with the RND-RF and Combine-RF.
This illustrates CTRF's ability to transfer knowledge from one domain to a similar but distinct domain with unstable factor (in this case, an ad's position). 

We calculate the probability of including the ``position'', which is a known spuriously correlated factor by design, in the top 5 factors ranking by feature importance \cite{genuer2010variable} evaluated on the training dataset. As shown in the right panel of Figure \ref{fig:auction_compare}, the random forest learned on the \rdata~ (RND-RF,CTRF are identical) has a lower probability of identifying the unstable or confounding factor as important predictors, compared with the one utilizing the \ldata~ (CNT-RF, Combine-RF). This demonstrates that the first stage of structure learning or the decision boundary on \rdata~ can reduce spurious correlation. This also validates utilizing the large amount of the \ldata~ to calibrate the parameters in the structure or trees in the second stage as the prediction does not rely on the unstable factor.

\section{Experiments on Real-World Data}
\label{sec:real-world}
In this section, we present experimental results in the real-world application with data collected from a sponsored search search platform (Bing Ads). 
First, we discuss how \rdata~ is collected from real traffic. Next, we demonstrate the robustness of CTRF-trained click models against the distribution shifts. Finally, we show that CTRF-enabled holistic counterfactual policy estimation improves global marketplace optimization problem real business scenarios.
\subsection{Randomized Experiment (\rdata)}
Randomized data (\rdata) collection is very important step to create CTRF since training requires \rdata~ to learn the structure of trees. In order to collect \rdata, we used existing randomization policy on paid search engine which is triggered less than \%1 of the live traffic. The existing randomization policy is triggered in typical sponsored search requests and there is no difference between randomized and mainstream traffic in terms of user and advertiser selection. For a given paid search request, if randomization is enabled, special uniform randomization policy is triggered. In this uniform randomization policy, all choices that depend on models are completely randomized. In particular, the ads are randomly permuted and the page layout (where ads are shown on the page) is chosen at random from the feasible layouts. The user cost (due to lower relevance) of such randomization is very high and consequently, limits the trigger rate for the randomized policy.

\subsection{Robustness to Real-World Data Shifts}
We train the user click model on the data collected from the mainstream traffic and randomized traffic in the search engine, corresponding to the \ldata~ and \rdata~ respectively. We validate the proposed method on an exploration traffic with some radical experiments (layout template change,  for example), which is the testing data with covariate shifts. We only compare the method with CNT-RF, Combine-RF and Oracle-RF, which trains a random forest on the testing data. The last one cannot be implemented in practice yet it serves to illustrate the capacity of the random forest method. We fix the total training size to be approximately 1 million with each method \footnote{The ratio of \rdata~ and \ldata~ is about 1:7, after down-sampling on the \ldata. The proportion of \rdata~ is upweighted for fair comparison. Otherwise, the performance of CNT-RF and Combined-RF will be very close.} and include the same feature set from production for a fair comparison. We focus on three metrics of interests: AUC (area under curve), RIG (Relative Information Gain) and cumulative prediction bias\footnote{Relative information gain is defined as the $\textup{RIG}=(H(\bar{y})+L)/H(\bar{y})$, $L$ is the log loss produced by the model and $H(p)=-p\textup{log}(p)-(1-p)\textup{log}(1-p)$ is the entropy function. Higher value indicates better performance.}.

\begin{table}[h]
  \caption{Performance comparison for different random forest based model, evaluated on some exploration flights with radical policy changes}
  \label{tab:ad-results-table}
  \centering
  \begin{tabular}{lccc}
    \toprule
   Methods    & AUC    & RIG & Cumulative Bias \\ 
    \midrule
   CNT-RF & 0.9273
  & 0.4424   & 3.87\% \\
   Combine-RF   & 0.9282& 0.4460 & 3.39\% \\
   CTRF     &  \textbf{0.9285}  & \textbf{0.4477} & \textbf{2.90}\%\\
   \hline
   Oracle-RF &  0.9287&  0.4484 & 0.58\%\\
    \bottomrule
  \end{tabular}
\end{table}

Table \ref{tab:ad-results-table} shows that CTRF achieves the best performance among all the random forest candidates\footnote{We omit the standard error here for brevity and the reported difference here is considered as ``significant'' in practical application.}. As for AUC and RIG, The CTRF shows a slightly better performance than other random forest candidates and is very close to Oracle-RF, which indicates its nearly-optimal prediction performance. In terms of the bias, although with a gap with the Oracle-RF, the CTRF reduces the cumulative bias for click rate prediction to a non-negligible degree, which is very essential to the publishers in decision making. As we are evaluating all the performance on a part of the traffic performing some radical changes, the results demonstrate that the CTRF improves the robustness of user click model in terms of prediction power.

\subsection{End-to-end Marketplace Optimization}
In addition to the prediction power of the model, we also evaluate how the usage of CTRF can advance the decision making procedure in real business optimization at Bing Ads.
\subsubsection{Marketplace Optimization in a Nutshell}
The goal of Marketplace Optimization for sponsored search is to find optimal operating points for each component of the search engine given all marketplace constraints. Marketplace optimization is very different from optimizing certain objective functions with a given machine learning model. While model training focuses on reducing prediction error for unobserved data, Ads Marketplace Optimization focuses on improving global objectives like total clicks, revenue when new machine learning model is used as part of a bigger system. Due to data distribution shifts between components of a larger system, a locally optimized click model does not necessarily give best performances for global metrics. Therefore, whole components of the system may need to be tuned together by using more holistic approaches like A/B testing \cite{xu2015infrastructure} or similar. 

\subsubsection{Experimental Data Selection and Simulation Setup}

Robust click prediction plays a very crucial role in improving holistic Ads Marketplace Optimizer like an open box simulator~\cite{bayir2019genie} which can easily have biased estimations due to data distribution shifts in counterfactual environments. In our problem context, we integrate CTRF to an open box offline simulator and show that a new simulator with CTRF will give better results for offline policy estimation scenarios when data distribution shift is significant.

For experimental runs, we use an open box simulator with two versions of random forest, CTRF and CNT-RF (typical RF used), along with the generalized linear Probit model \cite{graepel2010web} for click prediction. Then, we run offline counterfactual policy estimation jobs with modified inputs over logs collected from real traffic. Finally, we compare predictions for marketplace level click metrics with different models against A/B testing by using same production data that is collected from A/B testing experiment. 

To select experimental data, we checked the counterfactual vs factual feature distribution similarity of multiple real tuning scenarios in search engine traffic. We applied Jensen-Shannon (JS) divergence to compute the similarity of two distributions. Based on this distance metric, we selected 2 tuning use cases out of 10 candidate cases with significantly higher distribution shift, which fits the proposed approach. First use case belongs to capacity change for Text Ads blocks. Second use belongs to page layout change. This also demonstrates that drastic policy changes are common in online advertisement tuning tasks. Details on this procedure are included in the Supplementary Material.

\subsubsection{Experiments on Real Case Studies}

In the first case, the capacity of the particular ad block that contains Textual Ads was increased on the traffic in May 2019 for 10 days time period during A/B testing. The change was expected to increase both overall click yield and click yield on textual ad slice for target ad block. For simulator runs, we used 4.6 million samples from control traffic (\ldata) and 100K samples from the randomized traffic (\rdata) that belongs to 3 weeks time period before end date of A/B testing. The randomized traffic corresponds to page view requests where the mechanisms in online system are randomized, as described in Section 5.1.

In simulator runs with CTRF, we train the forest and tree structures from \rdata~ and combine the \ldata~ and \rdata~ to calibrate the leaves of trees in the forest. Each simulation job uses its trained model to score counterfactual page views that generated from replying control traffic logs in open box manner with the suggested input modification (capacity change of ad block). Table \ref{tab:ad-capacity-change} presents the comparison of an open box simulator with generalized Probit model, with CTRF and the random forest trained on control traffic (CNT-RF) based on relative Click Yield delta error \footnote{Relative Click Yield delta error is defined as |$\Delta \textup{CY}_{\textup{Method}} - \Delta\textup{CY}_{\textup{AB}}|/ |\Delta\textup{CY}_{\textup{AB}}$|. $\Delta \textup{CY}_{\textup{Method}}$ is the predicted change in click rate by the model. $\Delta\textup{CY}_{\textup{AB}}$ is the actual change in A/B testing. } against A/B testing experiment that was active for 10 days in May 2019. To make a fair comparison, we use the same amount of training data for different variants of random forest models.  We observe that click yield deltas coming from simulator results with CTRF is significantly better than other approaches since results from CTRF enabled simulator are closer to A/B Testing results from real traffic.

\begin{table}[h]
  \caption{Performance comparison in two cases with radical changes}
  \label{tab:ad-capacity-change}
  \centering
  \begin{tabular}{lcc}
    \toprule
  Ad capacity change    & $\Delta$CY Error        & $\Delta$CY Error (Text Ads) \\
    \midrule
   Probit Model & 34.94\%  & 17.13\% \\
   CNT-RF & 12.11\%  & 9.96\% \\
   CTRF  & \textbf{2.07\%} & \textbf{8.76\%} \\
   \midrule
   \bottomrule 
  Layout change    & $\Delta$CY Error       & $\Delta$CY Error (Shopping Ads) \\
    \midrule
Probit Model & 35.48\%  & 45.08\% \\
CNT-RF & 58.06\%  & 34.92\% \\
CTRF & \textbf{22.58\%} & \textbf{13.38\%} \\
    \bottomrule
  \end{tabular}
\end{table}

In the second scenario, the layout of product shopping ads was significantly updated in May 2019 for a week time period during A/B testing. The change was expected to increase both overall click yield and click yield on product shopping ads slice for target ad block. In this experiment, we used 15M samples from the control traffic in A/B testing and the same randomized traffic in the previous experiment. The bottom part in Table~\ref{tab:ad-capacity-change}
presents the comparison of different model-based simulators in the relative error against the A/B testing experiment that was active for a week in May 2019.
Since the modification for the second experiment yielded a radical shift in feature distribution of product shopping Ads. The difference with CTRF enabled simulator vs other approaches is more prominent. Thus, open box simulator with CTRF also outperforms other approaches in this scenario.

\section{Discussion and Conclusions}

We present a novel method, causal transfer random forest, to combine limited randomized data (\rdata) and large scale logged data (\ldata) in the learning problem. We propose to learn the tree structure or the decision boundary with the \rdata~ and calibrate the leaf value of each tree with the whole data (\rdata~ and \ldata). This approach overcomes the spurious correlation in \ldata~ and the limitations on sample size for the \rdata~ to provide robustness against covariate shifts. We evaluate the proposed model in the extensive synthetic data experiments and implement it in Bing ads system to train the user click model. The empirical results demonstrate its advantage over other baselines against the radical policy changes and robustness in real-world prediction tasks. For future work, there are some important research questions to explore, such as a better understanding of the relative importance of the \rdata~ versus the \ldata, how much \rdata~ is needed and how this quantity related to the degree of distributional shift.








\bibliography{TransferRF.bib}
\bibliographystyle{ACM-Reference-Format}

\end{document}


\newpage

\appendix

\section*{A. Note on Reproducibility}

For the research presented in this paper, We implemented two versions of our algorithm, including a version tailored for a large-scale distributed analytics platform for the real-world experiments, and another version suitable for use in a single-machine python environment. We enclose the code for the single-machine implementation of our algorithm along with the submission.

For our reported synthetic data experiments, we provide a detailed description of the data generating process and the implementation of all the methods in the experiments. For the real-world experiments on a large-scale distributed analytics platform, we describe the details of the experiments to enhance the understanding of the role of our proposed method.

The environment we use is:
\begin{itemize}
    \item \textbf{Python 2.8.1} 
    \item \textbf{scikit-learn=0.22.1}
    \item \textbf{numpy=1.18.1}
    \item \textbf{pandas=1.0.0}
\end{itemize}

Codebase to reproduce the experimental results:
\begin{itemize}
	\item \textbf{run\_simulation.py}:Main script for the simulation in Section 4.1
	\item \textbf{run\_auction.py}: Main script for the simulated auction in Section 4.2.
	\item \textbf{ctrf}: Module contains the CTRF implementations.
\end{itemize} 

We also provide two example bash scripts \textbf{simulated\_data\_varying.sh} for running the simulated experiments in Section 4.1 and \textbf{simulated\_auction\_reverse\_varing,sh.} for simulated auction in Section 4.2

The implementation details for each algorithm:
\begin{itemize}
    \item LR: The \textbf{LogisticRegression} class in \textbf{sklearn.linear\_model}, with $L_{2}$ penalty.
    \item GBDT: \textbf{GradientBoostingClassifier } class\\ in \textbf{sklearn.ensemble}.
    \item CNT-RF/RND-RF/Combine-RF: \textbf{RandomForestClassifier} class in \textbf{sklearn.ensemble}, trained on corresponding data.
    \item CTRF: Python module provided in codebase.
\end{itemize}

\noindent For methods adjusted with sampling weights (LR-IPW,GBDT-IPW), we first estimate the ratio of density with a logistic regression to classify training data and testing data \cite{bickel2009discriminative}. We calculate the IPW weights $w_{i}=p^{\textup{test}}_{i}/(1-p^{\textup{test}}_{i})$, where $p^{\textup{test}}_{i}$ is the predicted probability for unit $i$ belongs to testing data. We feed the IPW weights when fitting model with, \textbf{model.fit(X,y,sample\_weight=weights)}, where\textbf{ weights }is the IPW weights.

For the hyperparameters in the random forest model, we set the number of trees to be $50$, $0.3$ feature subsampling rate and the max number of nodes to be 100 (same for GBDT model). 
\section*{B. Details on Experiments}

\subsection*{B.1 Details on Synthetic Auction}
We enumerate the steps for generating the synthetic auction data. 
\begin{itemize}
    \item Step 1: We utilize the \textbf{scikit-learn.make$\_$classification} function to generate synthetic relevance data $(x,y)$. Each data point corresponds to one ad to be shown.
    \item Step 2: We fit a random forest model to the data to calculate a relevance score/probability of being click $p$ for each ad.
    \item Step 3: We run a simulated auction based on the relevance score with some additional noise $p'$. Each auction determines the ad's layout on one page. In each auction, 20 ads are being considered to compete for the position in the layout with at most 5 slots. Notice that the relevance reserve serves as a filter to determine whether the ad can join the auction.
    \item Step 4: we assign position based on the $p'$ in the auction with high relevance assigned to the top position.
    \item Step 5: We generate click based on true relevance score $p$ with Bernoulli trials and randomly pick up one ad to click if the user would click multiple ads on the same page. 
\end{itemize}

\noindent For randomized data, we skip the auction stage and simply randomly pick some ads to show on the page. We run 10000 auctions for the randomized data and 25000 auctions on the log data. The final sample size ratio between randomized and log data is approximately 1:5.

\subsection*{B.2 Details on End-to-End Optimization Task}
We also provide a detailed description on how we calculate the degree of feature shifts in real-world task and how we pick up the optimization tasks.

In order to compute distribution shift between two different environments, we use discrete bins to represent each feature as a multinomial distribution similar to approach described in ~\cite{bayir2019genie}. After that, we applied Jensen-Shannon (JS) divergence metric to compute the similarity of two multinomial distribution for the same feature in counterfactual vs factual environment. We select the typical cases with lower similarity to demonstrate the use of the the proposed method. The JS Divergence of two probability distribution P and Q are given below:

\begin{align}\label{eq:JS}
	JS(P||Q) = \frac{1}{2}D_{KL}(P||M) + \frac{1}{2}D_{KL}(Q||M),
	M = \frac{1}{2}(P + Q)
\end{align}

\noindent JS Divergence is the symmetric version of Kullback–Leibler divergence which can be computed as below for a given multinomial distribution with $k$ different bins.

\begin{align}\label{eq:KL-Divergence}
	D_{KL}(P || Q) = \sum_{i}^{k}(P(i)\textbf{ }ln(\frac{P(i)}{Q(i)}))
\end{align}

\noindent Once the JS divergence of each feature is computed based on counterfactual ($P$) vs factual ($P^{*}$) feature distributions, the final distribution shift score (DS) over $N$ features is computed as root mean square of all JS divergence values across all features as follows:

\begin{align}\label{eq:DS-Metric}
	DS = \sqrt{\frac{1}{N}\sum_{i}^{N}\left[JS(P_{i}||P^{*}_{i})\right]^{2}}
\end{align}

\noindent The distribution shift (DS) scores for selected real use cases are given below in Table~\ref{tab:distribution-shift}. We calculate the feature shifts of 10 candidate task in total and compare the two tasks in the paper with other eight tasks. Based on the JS-divergence metrics, the two tasks we demonstrate in the paper serve as good examples for covariates shifts and drastic change in the mechanism.

\begin{table}[H]
  \caption{Comparison for distribution shifts}
  \label{tab:distribution-shift}
  \centering
  \begin{tabular}{lc}
    \toprule
  DS (Text Ads Case) & $10^{-2}$\\
  DS (Shopping Ads Case) & $5x10^{-3}$\\
  Average DS (Others) & $45x10^{-5}$\\
  STD of DS (Others) &  $35x10^{-5}$\\
    \bottomrule
  \end{tabular}
\end{table}

\section*{C. Proof for Theorems}
First, we give explicit description for the theorems in Section 3.3 which are from previous literature. The first theorem in \cite{rojas2018invariant} establishes the relationship between conditional invariant property and robust prediction

\begin{theorem}[Adversarial robustness]
\label{thm:invariant}
Suppose we have training data from various sources. $\{(x^{k}_{i},z^{k}_{i},y^{k}_{i})\}\sim \mathcal{P}^{k},k=1,2,\cdots,K$ and wish to make prediction on targeting $\{(x_{i}^{K+1},z^{K+1}_{i},y_{i}^{K+1})\}\sim \mathcal{P}^{K+1}$. Assume there exists a unique subset of features $S^{\ast}$ such that: $y_{i}^{k}|S^{\ast k}_{i}\stackrel{d}{=}y_{i}^{k'}|S^{\ast k'}_{i},k\neq k'\in\{1,2,\cdots K+1\}$ (conditional invariant property). Then:\\
\begin{eqnarray}
E_{\mathcal{P}^{1,\cdots,K}}(y_{i}|S_{i}^{\ast})=\textup{argmin}_{f}\textup{sup}_{(x_{i},z_{i},y_{i})\sim \mathcal{P}}E||f(x_{i},z_{i}),y_{i}||^{2},
\end{eqnarray}
where $\mathcal{P}$ is the family of distributions of $(x_{i},z_{i},y_{i})$  satisfying the invariant property. $\mathcal{P}^{1,\cdots,K}$ is the distribution pooling $\mathcal{P}^{1},\mathcal{P}^{2}\cdots,\mathcal{P}^{k}$ together.\\
\end{theorem}

The second theorem from \cite{peters2016causal,rojas2018invariant} states relationship between conditional invariant property and causality.

\begin{theorem}[Relationship to causality]
\label{thm:direct_cause}
If we further assume $(x_{i},z_{i},y_{i})$ can be expressed with a direct acyclic graph (DAG) or structural equation model (SEM). Namely, let $c_{i}=(x_{i},z_{i})$, $c_{i}^{j}=h_{j}(c^{\textbf{PA}_{j}}_{i},e^{j}_{i})$, $y_{i}=h_{y}(c^{\textbf{PA}_{Y}}_{i},e_{i})$. Then we have
$S_{i}^{\ast}=c_{i}^{\textbf{PA}_{Y}}$,
where $c^{\textbf{PA}_{j}}$ denotes the parents for $c_{j}$, $c^{\textbf{PA}_{Y}}$ denotes the parents for $y$, $e_{i}^{j},e_{i}$ are the noises, $h_{j}(\cdot,\cdot)$ and $h_{y}(\cdot,\cdot)$ are deterministic functions.
\end{theorem}

Now we prove the theorem in the main text to validate the use of \rdata,
\textit{Proof:} Assuming certain regularity conditions, such as the integrals are well-defined, suppose the model trained can converge to the conditional mean,
\begin{eqnarray}
E(y_{i}|x_{i},z_{i})&\rightarrow_{p} &\int_\mathcal{Y} yp(y|x,z)dy=\int_\mathcal{Y}y\frac{p(y,x,z)}{p(x,z)}dy
\end{eqnarray}
Furthermore, under randomization conditions, we have,
\begin{eqnarray}
\int_\mathcal{Y}y\frac{p(y,x,z)}{p(x,z)}dy&=&\int_\mathcal{Y}y\frac{p(y,x,z)}{p(x_{i}^{1})p(x_{i}^{2}\cdots p(x_{i}^{p})p(z_{i}^{1})\cdots p(z_{i}^{p'}))}dy\\
&=&
\int_\mathcal{Y}y\frac{p(y|c_{i}^{PA_{Y}})p(x_{i}^{1})p(x_{i}^{2})\cdots p(x_{i}^{p})p(z_{i}^{1})\cdots p(z_{i}^{p'})}{p(x_{i}^{1})p(x_{i}^{2})\cdots p(x_{i}^{p})p(z_{i}^{1})\cdots p(z_{i}^{p'})}dy\\
&=&
\int_\mathcal{Y}y\frac{p(y(\textup{do}(c_{i}^{PA^{Y}}))p(x_{i}^{1})p(x_{i}^{2})\cdots p(x_{i}^{p})p(z_{i}^{1})\cdots p(z_{i}^{p'})}{p(x_{i}^{1})p(x_{i}^{2})\cdots p(x_{i}^{p})p(z_{i}^{1})\cdots p(z_{i}^{p'})}dy\\
&=&E(y_{i}|c_{i}^{PA_{Y}})=E(y_{i}|S_{i}^{\ast})
\end{eqnarray}

\bibliography{TransferRF.bib}
\bibliographystyle{ACM-Reference-Format}